\documentclass[runningheads]{llncs}
\usepackage[T1]{fontenc}
\usepackage{graphicx,verbatim}
\usepackage{graphicx}
\usepackage{booktabs}
\usepackage{subcaption}
\usepackage{multirow}
\usepackage{siunitx} 
\usepackage{tikz}
\usetikzlibrary{arrows.meta, positioning, shapes.geometric}
\usepackage{todonotes}
\usepackage{hyperref}

\begin{document}
%
\title{Understanding Model Behavior in \\ Monocular Polyp Sizing}
%

\author{Xinqi Xiong$^{1}$, Andrea Dunn Beltran$^{1}$, Junmyeong Choi$^{1}$, Sarah K. McGill$^{1}$, \\Marc Niethammer$^{2}$, Roni Sengupta$^{1}$}  
\authorrunning{X. Xiong et al.}
\institute{$^{1}$University of North Carolina at Chapel Hill  
    \quad $^{2}$University of California, San Diego}
  
\maketitle              

\begin{abstract}

Accurate polyp size stratification guides surveillance decisions, with lesions larger than 5\,mm typically requiring closer follow-up.
However, monocular colonoscopy lacks a reliable metric reference. We present a diagnostic audit of binary polyp size classification ($\le$5\,mm vs. $>$5\,mm) across multiple public multi-center datasets, model families, and patient-stratified cross-validation.
Across architectures and input modalities, including RGB appearance, relative depth, and photometry, model performance is moderately consistent, suggesting reliance on cues correlated with examination behavior rather than true metric scales.
By providing ground-truth scale at varying granularities, we quantify the potential improvement from perfect scale information and show that current depth estimation and global calibration offer limited gains.
We further demonstrate that segmentation errors under distribution shift eliminate most of this potential, with oracle scale under predicted masks recovering only baseline performance.
These results highlight metric scale and mask robustness as two independent bottlenecks and provide reusable evaluation tools—such as oracle scale ladders, shortcut partitions, and mask substitution—for auditing future polyp sizing pipelines. Our code is publicly accessible at \href{https://github.com/anaxqx/polyp-sizing-audit}{https://github.com/anaxqx/polyp-sizing-audit}.


 \end{abstract}
    
\section{Introduction}
\label{sec:intro}
Colorectal cancer is the second leading cause of cancer death~\cite{siegel2024cancer}. The clinical management of colorectal polyps is risk-stratified by size: diminutive polyps ($\le 5$\ mm) often allow for a 7–10 year surveillance interval if only 1–2 are found, while small polyps (6–9 mm) may require a 3–5 year follow-up depending on the total count and pathology. Large polyps ($\ge 10$\ mm) are considered advanced findings that necessitate immediate resection and a mandatory 3-year surveillance interval~\cite{gupta2020recommendations}. However, in current clinical practice, physicians inspect the polyp visually and qualitatively estimate the size, often leading to a 40\% error rate~\cite{cheloff2022accuracy}. 
This inaccuracy introduces systematic bias and directly alters the surveillance intervals for approximately 10\% of patients~\cite{chaptini2014variation}.

Automated polyp sizing could reduce inter-observer variability and support consistent surveillance decisions~\cite{polypsize}.
However, monocular endoscopy presents a geometric ambiguity: without a metric reference, a 3\,mm polyp at a 10\,mm distance is visually indistinguishable from a 6\,mm polyp at 20\,mm~\cite{hewett2022measurement}. 
This scale ambiguity is further compounded by clinical examination behavior. Endoscopists naturally move the camera closer to diminutive polyps to confirm their status while evaluating larger lesions from a wider field of view. 
This can induce a spurious correlation between apparent scale and the size label in naturalistic training data.
Furthermore, due to variability in endoscopy devices and imaging protocols in different clinical centers, the models trained on one dataset often struggle to generalize.

Prior automated approaches fall into three categories: explicit 3D reconstruction methods~\cite{iranzo2025endometric,du2024polyp,ruano2024estimating}, reference-based methods~\cite{sudarevic2023artificial}, and end-to-end learned models~\cite{itoh2018towards,itoh2021binary,polypsize,liupolypsense3d,roffo2024feature}. 
3D reconstruction methods require recovering full 3D geometry from monocular endoscopy, a problem that remains challenging in feature-poor, deformable colon environments~\cite{bonilla2024gaussian}. 
In effect, they attempt to solve a harder problem as a prerequisite for a simpler one. 
Reference-based methods that use instruments such as forceps or water jets as scale anchors report low measurement error in controlled settings~\cite{sudarevic2023artificial,liupolypsense3d}, but require the reference object to be present during the procedure, limiting scalability. 
End-to-end learned models report high accuracy on single-center datasets~\cite{itoh2018towards,itoh2021binary}, but most implementations are not publicly available, and it remains unclear whether their reported gains reflect genuine recovery of metric scale or reliance on procedural shortcuts. 
We replicate representative approaches~\cite{itoh2021binary,polypsize,roffo2024feature}, and find that models across input modalities converge to the same performance ceiling, with diagnostic evidence pointing to shortcut reliance rather than geometric reasoning~\cite{geirhos2020shortcut,degrave2021ai,castro2020causality}. 


Despite growing interest in automated polyp sizing, the field lacks a controlled analysis of what limits current models and where future effort should be directed. We address this gap with a diagnostic benchmark for monocular polyp size classification ($\le$ 5\,mm vs.$>$5\,mm), evaluated across two public multi-center datasets~\cite{realcolon,sunseg}, multiple model families, and patient-stratified cross-validation. 

In this paper, we identify three primary factors that severely limit progress in monocular polyp size estimation. 
First, collecting large-scale, size-annotated multi-center data remains costly and logistically challenging, restricting the diversity of training distributions and resulting in poor generalization~\cite{realcolon,polypsize,liupolypsense3d,sunseg}. 
Second, recent foundation models for metric depth estimation~\cite{zoedepth,depthpro} show strong zero-shot performance in natural scenes but fail under specular, textureless, and optically distorted endoscopic conditions. 
Third, although polyp segmentation models report high in-domain accuracy~\cite{polypPVT}, it remains unclear whether this performance transfers to downstream tasks such as sizing under domain shift.

Our contributions are:
\begin{enumerate}
 \item We show that models across architectures and input modalities (appearance, relative depth, photometry) reach a similar performance plateau (Macro-F1 $\approx$ 0.75). We provide diagnostic evidence that this ceiling reflects shortcut reliance on examination behavior rather than genuine geometric reasoning.
 \item We introduce oracle scale probes at multiple granularities (per-frame, per-polyp, global) to isolate the performance headroom attributable to scale ambiguity (+16.1\,pp), and show that current metric depth estimators do not recover this headroom in clinical data.
 \item We demonstrate that segmentation quality acts as an independent bottleneck: replacing ground-truth masks with predictions from a segmentation model not trained on the target datasets erases the gains from oracle scale (-15.3\,pp), indicating that in-domain segmentation benchmarks do not predict downstream sizing robustness under distribution shift.
 \end{enumerate}

\section{Problem Formulation}
\label{sec:framework}

We study monocular polyp sizing under a geometric scale ambiguity. Under a pinhole camera model, the apparent diameter $A$ (pixels) of a polyp with true diameter $S$ at distance $Z$ is
\begin{equation}
    A = \frac{f S}{Z},
    \label{eq:pinhole}
\end{equation}
where $f$ is the focal length. Since $A$ depends on both $S$ and $Z$, $S$ is not identifiable from monocular appearance without a metric reference.

Clinical examination behavior further complicates this setting. Endoscopists often move closer to small polyps and view larger lesions from farther away, which can couple $S$ and $Z$ through behavior $B$ ($S \rightarrow B \rightarrow Z$). As a result, the apparent scale $A$ and procedure context $B$ can correlate with the size label $S$, providing shortcuts that may remain stable across environments.

Motivated by this formulation, we audit monocular sizing via three independently controllable factors: data--model bias, metric scale, and mask quality. Each factor is evaluated by a targeted intervention in Sec.~\ref{sec:results_plateau}--\ref{sec:results_masks}.

\section{Experimental Setup}
\label{sec:setup}
\begin{table}[t]
\caption{Binary polyp size classification ($\le$5\,mm vs.\ $>$5\,mm). Results are mean$\pm$std over 5-fold polyp-stratified CV. (Scale regimes follow Sec.~\ref{sec:results_scale}: \textit{None} uses depth as-is; \textit{Global} fits one scale factor on the training set; \textit{Oracle-polyp} \& \textit{Oracle-frame} fit per-clip \& per-frame factors. Pred masks are from PolypPVT.)}
\label{tab:table1}
\centering
\resizebox{\columnwidth}{!}{%
\begin{tabular}{l l l l | c c}
\hline
\textbf{Model} & \textbf{Input} & \textbf{Scale} & \textbf{Mask} & \textbf{Macro-F1} & \textbf{>5\,mm Rec.} \\
\hline
\multicolumn{6}{c}{\textit{Data-model bias (cross-family comparison)}} \\
\hline
ResNet18 & RGB & --- & GT bbox & $0.751 \pm 0.058$ & $0.673 \pm 0.141$ \\
ViT-B & RGB & --- & GT bbox & $0.778 \pm 0.097$ & $0.664 \pm 0.275$ \\
CNN3 & Rel.\ depth~\cite{ppsnet} & None & GT bbox & $0.725 \pm 0.048$ & $0.733 \pm 0.099$ \\
MLP (w. geom) & Photometry & --- & --- & $0.752 \pm 0.055$ & $0.704 \pm 0.126$ \\
MLP (w.o. geom) & Photometry & --- & --- & $0.694 \pm 0.057$ & $0.616 \pm 0.133$ \\
ResNet18$+$V-REx & RGB & --- & GT bbox & $0.753 \pm 0.058$ & $0.663 \pm 0.140$ \\
\hline
\multicolumn{6}{c}{\textit{Metric scale ladder (CNN3; baseline Macro-F1=0.725)}} \\
\hline
CNN3 & Metric depth~\cite{liu2025metriccol} & None & GT bbox & $0.727 \pm 0.055$ & $0.751 \pm 0.094$ \\
CNN3 & Rel.\ depth~\cite{ppsnet} & Global & GT bbox & $0.716 \pm 0.057$ & $0.742 \pm 0.105$ \\
CNN3 & Rel.\ depth~\cite{ppsnet} & Oracle-polyp & GT bbox & $0.792 \pm 0.033$ & $0.778 \pm 0.104$ \\
CNN3 & Rel.\ depth~\cite{ppsnet} & Oracle-frame & GT bbox & \textbf{$0.886 \pm 0.028$} & $0.876 \pm 0.047$ \\
\hline
\multicolumn{6}{c}{\textit{Mask quality robustness (CNN3 baseline=0.725; ResNet18 baseline=0.751)}} \\
\hline
CNN3 & Rel.\ depth~\cite{ppsnet} & Oracle-frame & GT bbox $\times 0.8$ & $0.852 \pm 0.035$ & $0.831 \pm 0.055$ \\
CNN3 & Rel.\ depth~\cite{ppsnet} & Oracle-frame & GT bbox $\times 1.2$ & $0.859 \pm 0.027$ & $0.847 \pm 0.050$ \\
CNN3 & Rel.\ depth~\cite{ppsnet} & None & Pred mask & $0.572 \pm 0.025$ & $0.559 \pm 0.076$ \\
CNN3 & Rel.\ depth~\cite{ppsnet} & Oracle-frame & Pred mask & $0.736 \pm 0.030$ & $0.735 \pm 0.053$ \\
ResNet18 & RGB & --- & GT bbox $\times 0.8$ & $0.758 \pm 0.072$ & $0.659 \pm 0.161$ \\
ResNet18 & RGB & --- & GT bbox $\times 1.2$ & $0.753 \pm 0.058$ & $0.644 \pm 0.163$ \\
ResNet18 & RGB & --- & Pred mask & $0.643 \pm 0.070$ & $0.505 \pm 0.175$ \\
\hline
\end{tabular}%
}
\vspace{1mm}
\end{table}

We establish an experimental protocol that isolates the contribution of each input modality from architectural choice. This section describes the data, evaluation, and diagnostic probes.

\subsection{Datasets and Evaluation}

We aggregate two public datasets with endoscopist-reported visual size estimates: Real-Colon~\cite{realcolon} (351,264 frames, 60 patients, 132 polyps) and SUN-SEG~\cite{sunseg} (49,136 frames, 99 patients, 100 polyps). The combined dataset contains 400,400 frames from 159 patients but only 232 unique polyps (of which 147 are $\le$ 5 \,mm and 85 are > 5\,mm). 
Given the high temporal correlation between adjacent video frames, the effective sample size is defined by the number of unique polyps rather than the total frame count. 

All experiments use 5-fold cross-validation stratified by patient: all polyps from the same patient appear exclusively in the same fold to prevent data leakage. Metrics (Macro-F1 and >5\,mm Recall) are reported as mean $\pm$
 standard deviation across folds. We focus on the 5\,mm threshold, which corresponds to the clinically actionable boundary between diminutive and small polyps~\cite{gupta2020recommendations}. The 10\,mm threshold is excluded because over 85\% of polyps fall below it, creating a class imbalance that precludes reliable evaluation.

\subsection{Diagnostic Probes and Interventions}

We design three lightweight probes, each restricted to a single input modality: (A) a ResNet18 on masked RGB, (B) the depth-based sizing three-block CNN of Itoh et al.~\cite{itoh2021binary} on relative depth maps from PPSNet~\cite{ppsnet}, and (C) a two-layer MLP on 51 handcrafted geometric and photometric features. A ViT-B on masked RGB is included to assess the effect of higher capacity. All models share a unified training pipeline: AdamW($\eta=10^{-4}$, $\lambda=10^{-4}$), 30 epochs with cosine annealing, and class-weighted cross-entropy.

To disentangle metric scale from segmentation quality, we apply controlled interventions to the CNN3 depth probe. 
Scale correction multiplies relative depth by a metric factor derived from Eq.~\ref{eq:pinhole}, an operation that is only meaningful for depth-based input. 
For the RGB probe, there is no natural mechanism to inject scale information, and the MLP features do not include raw depth values.
For scale, we estimate scale correction factors at three granularities using Eq.~\ref{eq:pinhole}: per-frame (\textit{Oracle-frame}), per-polyp clip (\textit{Oracle-polyp}), and a single global factor from the training set (\textit{Global}). 
We also evaluate zero-shot metric depth from MetricCol~\cite{liu2025metriccol}. 
For masks, we replace ground-truth bounding boxes with PolypPVT~\cite{polypPVT} predictions and evaluate $\pm$20\% bounding-box perturbations.

\section{Results and Discussion}
\label{sec:results_discussion}

\subsection{Data--Model Bias}
\label{sec:results_plateau}

Table~\ref{tab:table1} (top) summarizes performance across probes. 
Despite spanning three input modalities, all models fall within a narrow band near Macro-F1 $\approx 0.75$ (ResNet18: 0.751; CNN3: 0.725; MLP: 0.752). 
Increasing capacity yields no reliable gain.
ViT-B on masked RGB reaches 0.778 but shows high cross-fold variance in $>5$\,mm recall ($0.664 \pm 0.275$), consistent with overfitting at this sample size. 
Similarly, invariant training (V-REx~\cite{pmlr-v139-krueger21a}) does not change the plateau (+0.002), reflecting directionally similar confounds across both datasets~\cite{arjovsky2019invariant}

We next audit the predictive signals behind this saturation. 
Mutual information (MI) on the MLP features indicates that the apparent size of the polyp in the image (\texttt{apparent\_area\_frac}, the ratio of bounding-box area to image area, MI$=0.195$) is more informative than photometric features of background illumination (MI$=0.043$). 
Removing geometric features reduces Macro-F1 by 5.6\,pp (0.752$\to$0.694), confirming that models rely on apparent size, which conflates true scale $S$ with distance $Z$. 
Together, these diagnostics suggest that the observed plateau is driven by missing metric information in the input, rather than insufficient capacity. 


\paragraph{Implications.}
Increasing training data diversity is the most direct way to reduce procedure-linked cues. 
However, collecting large-scale, size-annotated multi-center data is costly, and our effective sample size of 232 polyps remains modest for deep learning models. 
The cross-fold instability of higher-capacity controls is consistent with sensitivity to procedure-specific signals and a sign of overfitting.

While synthetic data generation can provide the required variability in polyp size, camera pose, intrinsics, and lighting, models trained solely on synthetic data often suffer from sim-to-real domain gaps that limit their transfer to clinical environments~\cite{bonilla2024gaussian}.
Nevertheless, even with expanded data, the geometric ambiguity in Eq.~\ref{eq:pinhole} remains: without a metric reference, more data can likely shift the plateau but cannot remove the identifiability limit.
Next, we test whether providing a ground-truth metric scale improves performance.

\subsection{Metric Scale}
\label{sec:results_scale}

\begin{figure}[t]
    \centering
    \includegraphics[width=0.8\linewidth]{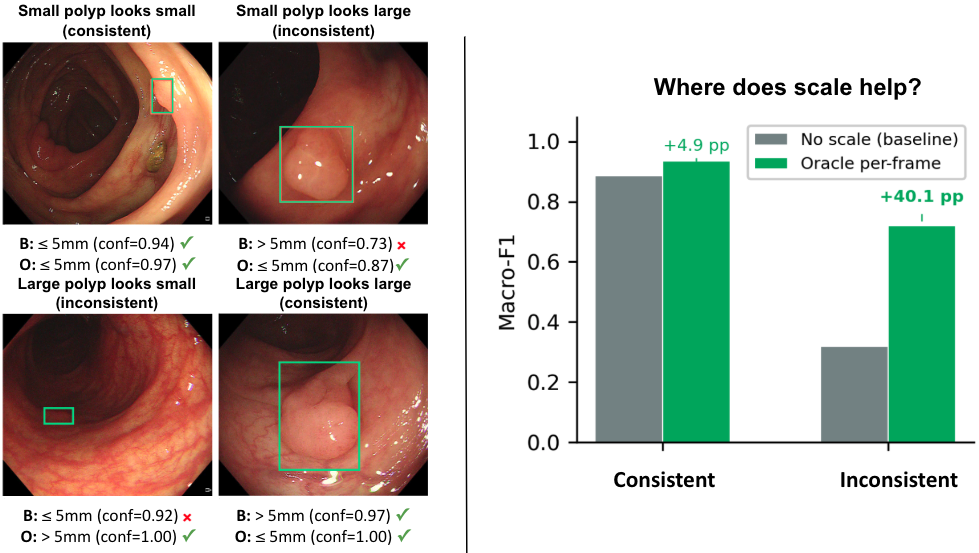}
    \caption{Effect of oracle per-frame scale on actual and apparent size consistent and inconsistent frames. (Left) Examples where apparent size matches (consistent) or contradicts (inconsistent) the true size, with predictions from the baseline (B) and oracle (O) models. (Right) Macro-F1 gains from oracle scale, concentrated in shortcut-inconsistent frames (+40.1 pp vs. +4.9 pp).}
    \label{fig:scale_shortcut}
\end{figure}

We conduct scale interventions on the CNN3 depth probe because scale correction operates on the depth map directly: the oracle multiplies relative depth by a metric factor, which is only meaningful for depth-based input. 
For the RGB probe, scale information has no natural injection point, and for the MLP, the handcrafted features do not include raw depth values.
Table~\ref{tab:table1} (middle) reports the effect of progressively resolving scale ambiguity. 
Oracle per-frame scale yields a +16.1\,pp gain (0.725 $\to$ 0.886). 
A per-polyp oracle scale provides a smaller +6.7\,pp gain (0.725 $\to$ 0.792), indicating that $Z$ varies substantially even within a single polyp clip, likely due to endoscope motion.

We also evaluate algorithmic alternatives. 
Replacing relative depth with metric depth, MetricCol~\cite{liu2025metriccol}, produces no measurable change (0.725 $\to$ 0.727, within fold variance). 
Global calibration, which fits a single scale factor on the training set, slightly degrades performance (0.725 $\to$ 0.716), indicating that scale variation is not captured by a population-level constant.

To localize where scale helps, we partition test frames into two groups based on whether the apparent size agrees with the true label (Fig.~\ref{fig:scale_shortcut}). 
In the \textit{consistent} group, polyps that appear large in the image are genuinely large, and polyps that appear small are genuinely small. 
In the \textit{inconsistent} group, apparent size contradicts the true physical size (e.g., a large polyp viewed from far away appears small). 
We split groups at the per-fold median of bounding-box area relative to image area.
The oracle gain concentrates in inconsistent frames (+40.1\,pp, 0.319 $\to$ 0.720) and is small in consistent frames (+4.9\,pp, 0.886 $\to$ 0.935). 
This pattern is consistent with metric scale resolving the $S/Z$ ambiguity when apparent size contradicts the true label.

\paragraph{Implications.}
In routine colonoscopy, camera-to-polyp distance varies continuously with scope advancement, angulation, and insufflation, yet few datasets~\cite{polypsize,liupolypsense3d,sunseg,endomapper} record camera intrinsics or any explicit metric reference. 
Recording intrinsics would benefit not only reconstruction methods but also physics-informed sizing models. 
However, knowing the focal length alone does not resolve ambiguity if the distance remains unknown. 
Reliable progress, therefore, likely requires either reference-based anchors (e.g., instrument tips) or calibrated sensing that provides a stable metric scale at the frame level.

Beyond deployment-time sensing, reference-based measurements during data collection could address a more fundamental issue: size labels in current datasets are endoscopist visual estimates and can be subject to rounding near clinical thresholds. 
Training and evaluation on tool-verified labels would reduce label noise at the decision boundary where sizing errors are most consequential.


Off-the-shelf metric depth models provide limited gains. 
Existing endoscopic metric depth methods~\cite{liu2025metriccol,li2025endostreamdepth} acquire scale from synthetic supervision and adapt to clinical data through self-supervised fine-tuning or data augmentation. 
In our evaluation, MetricCol~\cite{liu2025metriccol}, which follows this paradigm, yields no measurable improvement over relative depth (0.725 $\to$ 0.727), suggesting that synthetically derived scale does not transfer reliably across the heterogeneous intrinsics and acquisition conditions present in multi-center data. 
Test-time adaptation may improve robustness to low-level shifts such as lighting and tissue appearance, but examination behavior introduces a semantic-level confound that is unlikely to be resolved without explicit metric scale supervision. 
Accurate metric depth estimation for endoscopy could substantially reduce this bottleneck, but reliable methods that generalize across clinical settings remain limited.

The oracle results assume accurate foreground isolation using a ground-truth mask. 
We evaluate this dependency next.

\begin{figure}[t]
    \centering
    \includegraphics[width=\linewidth]{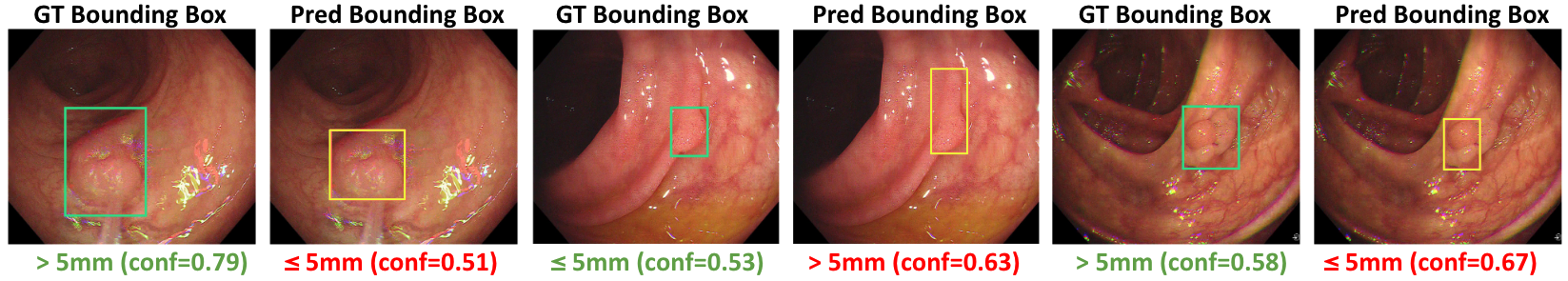}
    \caption{Effect of mask quality on sizing predictions. Each image pair shows the same polyp under ground-truth (GT) and predicted bounding boxes~\cite{polypPVT}, paired with model predictions from CNN3. Predicted masks introduce localization errors that change the apparent size seen by the model, directly altering the classification outcome (e.g., misclassifying a correctly classified polyp under GT bbox).
    }   
    \label{fig:mask_bottleneck}
\end{figure}

\subsection{Mask Quality}
\label{sec:results_masks}

Table~\ref{tab:table1} (bottom) evaluates the interaction between scale and segmentation quality. 
PolypPVT~\cite{polypPVT} reports mDice $> 0.85$ on curated benchmarks but achieves IoU $\approx 0.3$ on Real-Colon, where full-procedure artifacts are prevalent. 
Replacing ground-truth bounding boxes with these predicted masks reduces CNN3 performance by 15.3\,pp (0.725 $\to$ 0.572) (Fig.~\ref{fig:mask_bottleneck}). 
With oracle per-frame scale under-predicted masks, performance recovers only to 0.736, which is close to the unscaled baseline (0.725). 
This indicates that segmentation errors can absorb most of the oracle-scale gain.

Bounding-box perturbations ($\pm 20\%$) further show higher sensitivity for depth-based sizing than for the RGB probe (CNN3 + Oracle: $-3.4$\,pp; ResNet18: within variance). 
Overall, these results indicate that metric scale and segmentation quality constitute two independent bottlenecks for monocular polyp sizing.

\paragraph{Implications.}
PolypPVT reports strong overlap scores on curated benchmarks, but Real-Colon contains full-procedure artifacts that can shift mask quality. 
Small polyps may be poorly localized under specular highlights, larger lesions may be partially segmented due to complex texture, and instrument interaction can introduce partial occlusion. 
Such conditions are underrepresented in curated training data and affect the foreground region used by sizing pipelines. 
As a result, in-domain overlap metrics alone may not predict robustness for downstream size estimation under realistic clinical variation. 
Including sizing accuracy under predicted or perturbed masks can serve as a task-conditioned robustness check.

\subsection{Photometry Control}
\label{sec:results_photometry}

Finally, we test whether near-field illumination falloff can serve as a proxy for scale~\cite{liu2025metriccol,ppsnet}.
Under ideal conditions, background luminance $P$ would follow an inverse-square law ($P \approx k/Z^2$), allowing distance to be estimated from brightness.
In clinical endoscopy, Auto-Exposure Control continuously adjusts sensor gain and illumination intensity to maintain stable image brightness regardless of distance, which breaks this relationship and limits the usefulness of photometric features as a distance surrogate. 
Consistent with this, photometry-only correction ($A/P$) recovers only 17\% of the oracle-scale headroom (Macro-F1 0.760).


\section{Conclusion}
\label{sec:conclusion}

We present a diagnostic audit for monocular polyp size classification ($\le$5\,mm vs.\ $>$5\,mm) across two multi-center datasets with polyp-stratified cross-validation. 
Across architectures and input modalities, performance concentrates near Macro-F1 $\approx 0.75$, and diagnostics indicate reliance on stable cues correlated with examination behavior. 
We identify three independent factors that currently limit automatic polyp size estimation: the need for accurate per-frame metric scale, high-quality segmentation masks robust to domain shift, and large-scale multi-center datasets with reliable size annotations. 
When apparent size in the image agrees with true physical size, current models perform well; however, performance degrades substantially when examination behavior causes apparent size to contradict the true label, and oracle scale specifically resolves these cases (+40.1\,pp). 
These results provide reusable evaluation checks (oracle scale ladder, shortcut partitions, and mask substitution) and suggest that progress will depend on clinically grounded scale references, segmentation robustness under shift, and increased data diversity.


\section{Acknowledgement}

This work is supported by a National Institute of Health (NIH) project 
\#1R21EB035832 “Next-gen 3D Modeling of Endoscopy Videos”.

\bibliographystyle{splncs04}
\bibliography{ref.bib}
%




\end{document}